\documentclass[lettersize,journal]{IEEEtran}
\usepackage{amsmath,amsfonts}
\usepackage{algorithmic}
\usepackage{array}
\usepackage[caption=false,font=normalsize,labelfont=sf,textfont=sf]{subfig}
\usepackage{url}
\usepackage{verbatim}
\usepackage{graphicx}
\usepackage{times}
\usepackage{cite}
\usepackage{soul}
\usepackage{multirow}
\usepackage{multicol}
\usepackage{color}
\usepackage{balance}

\def\BibTeX{{\rm B\kern-.05em{\sc i\kern-.025em b}\kern-.08em
    T\kern-.1667em\lower.7ex\hbox{E}\kern-.125emX}}
\usepackage{balance}
\begin{document}
\title{Road Extraction with Satellite Images and Partial Road Maps}

\author{Qianxiong Xu, Cheng Long,~\IEEEmembership{Senior Member,~IEEE}, Liang Yu, Chen Zhang,~\IEEEmembership{Member,~IEEE}
\thanks{This work was supported in part by the Ministry of Education of Singapore under Academic Research Fund (Tier 2 Award MOE-T2EP20221-0013); and in part by Alibaba Group through Alibaba Innovative Research (AIR) Program and Alibaba-NTU Singapore Joint Research Institute (JRI), Nanyang Technological University, Singapore, under Award AN-GC-2020-006. Any opinions, findings and conclusions or recommendations expressed in this material are those of the author(s) and do not reflect the views of the Ministry of Education, Singapore. \textit{(Corresponding author: Cheng Long.)}}
\thanks{Qianxiong Xu and Cheng Long are with the School of Computer Science and Engineering, Nanyang Technological University, Singapore (e-mail: qianxion001@e.ntu.edu.sg; c.long@ntu.edu.sg).}
\thanks{Liang Yu is with Alibaba Cloud Computing, Alibaba, Hangzhou, China (e-mail: liangyu.yl@alibaba-inc.com).}
\thanks{Chen Zhang is with the Department of Computing, Hong Kong Polytechnic University, Hong Kong, SAR China (e-mail: c4zhang@comp.polyu.edu.hk).}
}

\markboth{Journal of \LaTeX\ Class Files,~Vol.~18, No.~9, September~2020}%
{How to Use the IEEEtran \LaTeX \ Templates}

\maketitle

\begin{abstract}
    Road extraction is a process of automatically generating road maps mainly from satellite images. Existing models all target to generate roads from the scratch despite that a large quantity of road maps, though incomplete, are publicly available (e.g. those from OpenStreetMap) and can help with road extraction. In this paper, we propose to conduct road extraction based on satellite images and partial road maps, which is new. We then propose a two-branch Partial to Complete Network (P2CNet) for the task, which has two prominent components: Gated Self-Attention Module (GSAM) and Missing Part (MP) loss. GSAM leverages a channel-wise self-attention module and a gate module to capture long-range semantics, filter out useless information, and better fuse the features from two branches. MP loss is derived from the partial road maps, trying to give more attention to the road pixels that do not exist in partial road maps. Extensive experiments are conducted to demonstrate the effectiveness of our model, e.g. P2CNet achieves state-of-the-art performance with the $IoU$ scores of 70.71\% and 75.52\%, respectively, on the SpaceNet and OSM datasets.
\end{abstract}


\section{Introduction}

\IEEEPARstart{R}{oad} maps are widely used in many applications including navigation, intelligent transportation, location-based services, urban design, etc. Conventionally, road maps are made manually via field surveys, which are costly and labor intensive and thus they could not be updated timely. In the last decade, extensive studies have been conducted for automatically extracting roads mainly from the satellite images, which is called \emph{road extraction}
~\cite{mnih2013machine,cheng2017automatic,mattyus2017deeproadmapper,zhou2018d,demir2018deepglobe,li2020topology}. 
Compared with conventional approaches, road extraction is significantly cheaper with little or no manual efforts and can be conducted frequently to reflect the updates such as newly-built roads and/or destroyed roads timely.

\begin{figure}[!t]
\centering
\includegraphics[width=0.8\linewidth]{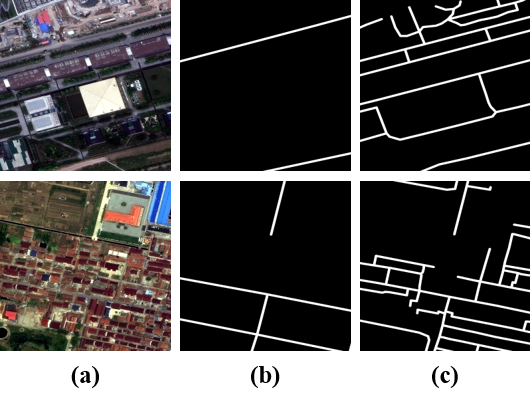}
\caption{Data samples. (a) Satellite image; (b) Road map (OSM); (c) Road map (SpaceNet).}
\label{osm_sample}
\end{figure}

Existing studies on road extraction all aim to conduct the task from an empty road map, i.e. they generate road maps from no road maps but satellite images only. While targeting this problem setting would reveal some scientific insights, e.g., how feasible it is to extract road maps from satellite images solely, it assumes a problem setting that is more difficult than necessary. In reality, a large quantity of road map data such as OpenStreetMap\footnote{\url{https://www.openstreetmap.org/}} (OSM), though not complete, has already been collected and made publicly available. 
For example, for Shanghai, the road maps available at OSM are about 54.19\% of those available at SpaceNet. Two samples of satellite images, partial road maps and complete road maps are shown in Figure~\ref{osm_sample}.
These partial road maps can greatly help with the road extraction task in the following aspects. 
First, they are already parts of the ground truth road maps, 
and thus there is no need to extract them again. Instead, we can focus more on the remaining parts.
Second, the satellite features of these available road maps can directly be obtained, 
with which the road extraction problem can be converted to one of finding features similar to these available features, which is much easier.

Therefore, we propose to conduct road extraction based on satellite images and partial road maps. There are mainly three possible options of utilizing the partial road maps. (1) Outputs fusion: we conduct road extraction without the partial road maps with an existing method, fuse the late-layer features with the partial road maps, and then use extra convolution layers to process the fused information. With this approach, the signal provided by the partial road maps would arrive a bit too late since it provides no guidance on the task of extracting roads from the satellite data and therefore it does not help to extract those roads that are missing. (2) Inputs fusion: we convert the partial road maps to images, fuse them and the satellite images as different channels, and then input the fused images to an existing method for extracting roads. Nevertheless, the model trained with this fused input would tend to reserve the roads in the partial road maps, but not to effectively extract those missing roads since the partial road maps are parts of the ground truth, i.e. the ability of extracting features would be weakened. (3) Features fusion: we extract features from both the satellite images and the partial road maps, fuse the extracted features, and then generate the road map from the fused features. This approach would avoid the issues of the first two options, i.e. the signal from partial road map comes early enough and it would not affect the extraction of satellite features directly, yet how to instantiate this design idea remains non-trivial.

In this paper, we design a features fusion approach. Specifically, we introduce a two-branch deep learning framework called Partial to Complete Network (P2CNet), where one branch takes satellite images as inputs and the other branch takes partial road maps as inputs. 
Each branch involves an encoder and a decoder, and for both of them we employ DeepLabV3+~\cite{chen2018encoder} with the backbone of ResNet-34 as the basic model. We develop modules for better utilizing and fusing the features outputted by the encoders. Specifically, a Gated Self-Attention Module (GSAM) is designed, which involves channel-wise attentions to better capture the long-dependency semantics from two branches and gate mechanisms to fuse the features from two branches in a complementary manner. 
Furthermore, apart from using the conventional binary cross entropy (BCE) loss and Dice loss, we introduce a Missing Part (MP) loss to further exploit the partial road maps by putting more attention on those road pixels that are missing in the partial road maps. With the MP loss, the roads that are missing would be better generated.

Extensive comparative experiments on models, fusion strategies, MP loss strategies, datasets and parameter settings are conducted and reported.
As our setting is new, we train some state-of-the-art models of semantic segmentation both in road extraction and road completion settings to provide better comparisons. 
The results show that P2CNet achieves the best results among others, as explained as follows.
First, the performance improvements of our method over the best baseline method (in terms of the commonly used metric IoU) is 1.25\% (0.7552 vs 0.7427) on OSM dataset and 0.52\% (0.7071 vs 0.7019) on SpaceNet dataset. In addition, the improvement is 1.85\% (0.8028 vs 0.7843) on OSM dataset for a topology-oriented metric APLS. 
Second, we emphasize that our method achieves the best results consistently across 6 metrics (including the topology-oriented one APLS), on 2 datasets of different qualities (OSM and SpaceNet), and among 11 methods. 
Third, our method is robust against different settings of partial road maps. For example, in the case that there are no partial maps (or they are purely black images), which reduces to the conventional setting of road extraction, our method is still slightly better than the state-of-the-art DeepLabV3+ (0.5297 vs 0.5277) on SpaceNet and DLinkNet (0.5243 vs 0.5240) on OSM for IoU.

The innovation of this paper is two-fold. First, it is an novel idea to conduct the road extraction task with some partial road maps as inputs since they are often available in practice and there is no good reason not to use them for road extraction. We note that the performance with partial maps is around 20\% better than without them for IoU on OSM dataset, from around 0.52 to around 0.75. With the improved performance, the road extraction task becomes more meaningful.
Second, we have made notable technical contributions for the fusion component and the loss component. For the fusion component, we designed (a) a fused self-attention mechanism based on \emph{affinity matrix multiplication}; and (b) a learnable gate module for deciding the confidence scores of pairs of feature maps from each branch such that the scores are computed in a complementary manner and adaptively adjusted based on data.
For the loss component, we developed a new function called \emph{missing part} (MP) loss to suit our new setting of using partial maps better. We combined this MP loss with the conventional binary cross entropy (BCE) loss and the Dice loss in the loss component. The experiments validate the benefits of the new MP loss.

In the remainder of the paper, we review the related work in Section~\ref{sec:related-work}, present our method, experiments, and experimental results in Section~\ref{sec:method}, Section~\ref{sec:experiments}, and Section~\ref{sec:results}, respectively. Finally, we conclude the paper in Section~\ref{sec:conclusion}.

\section{Related Work}
\label{sec:related-work}

\begin{figure*}[t]
\centering
\includegraphics[width=1\linewidth]{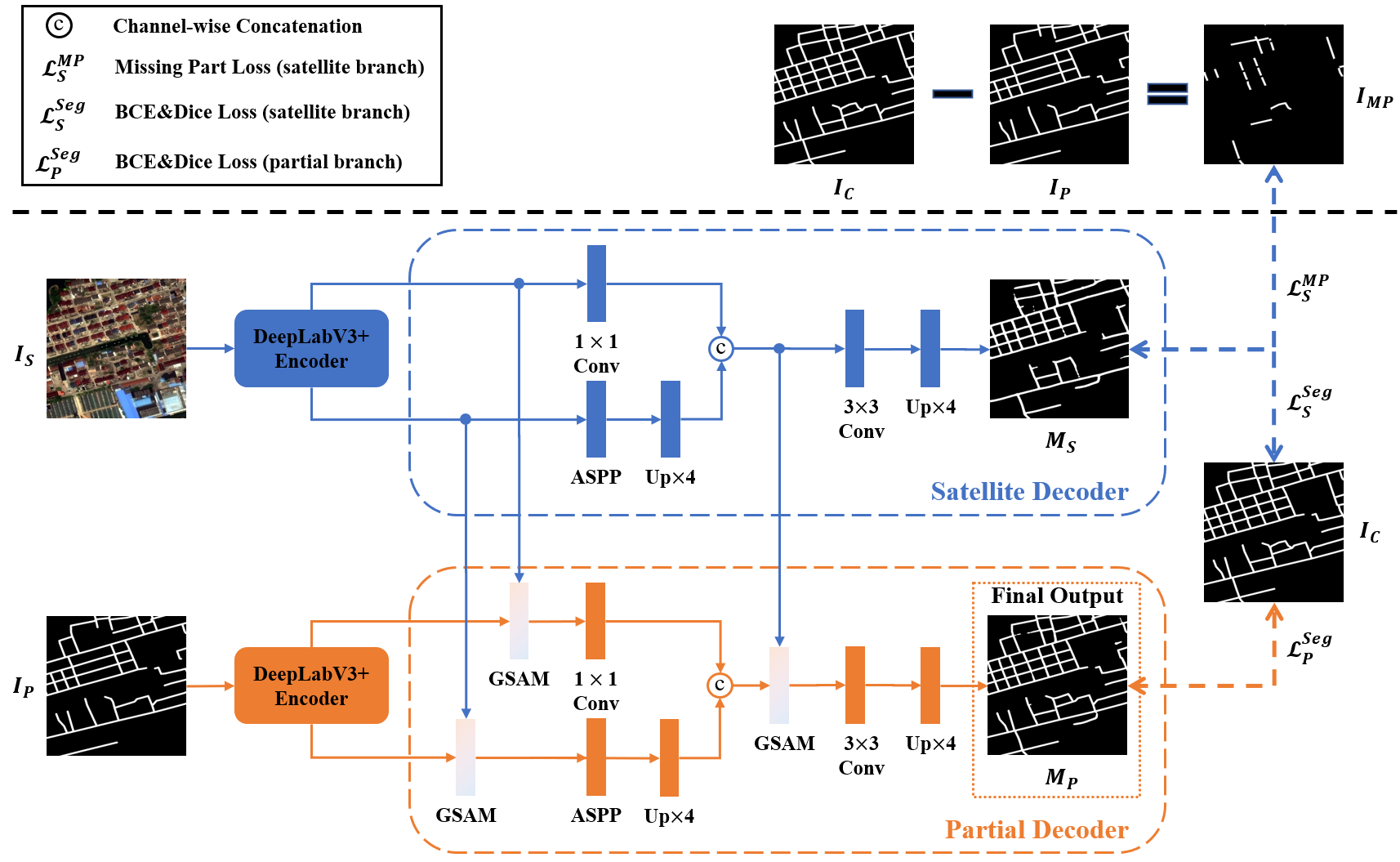}
\caption{Overview of the proposed P2CNet model. GSAMs are installed in the partial branch. There are also 4 GSAMs installed in the partial encoder, which is not explicitly displayed in the figure. The MP loss is applied to the satellite branch only.}
\label{p2cnet}
\end{figure*}

\noindent In this section, we briefly review some related studies about road extraction, data fusion and road inpainting.

\subsection{Road Extraction}
\noindent 
Researchers first adopted histogram, clustering and morphological techniques to extract roads~\cite{singh2013automatic,singh2014two}.
Then, Mnih et al.~\cite{mnih2013machine} built a patch-based deep convolutional neural network (DCNN) for road extraction.
\cite{zhong2016fully} then utilized FCN~\cite{long2015fully} for the task. 
Later, a variant of FCN named DeconvNet gradually became the main stream.
With the success of UNet~\cite{ronneberger2015u} in the field of medical image segmentation, Zhang et al.~\cite{zhang2018road} designed a deep residual UNet model to extract road features from remote sensing imagery data, which guarantees faster convergence speed during training and achieves good results.

Based on standard road extraction models,
some auxiliary tasks are adopted to obtain fine-grained features, which are then used to help extract the roads.
For example, Cheng et al.~\cite{cheng2017automatic} proposed a cascade model that is responsible for extracting road areas and road centerlines at the same time. 
Li et al.~\cite{li2020gated} combined road extraction with edge detection task, which was used to identify the boundary of roads, and obtained better results.
Moreover, \cite{mattyus2017deeproadmapper} used a model for initial road extraction first, then focused on mitigating the connectivity problem of the extracted roads, and improved the topology.
Some studies~\cite{abdollahi2020vnet,mosinska2018beyond,wei2017road} 
proposed novel loss functions to provide extra regulations and further improved the segmentation performance, e.g., \cite{abdollahi2020vnet} used dice loss to mitigate the class imbalance problem, and \cite{mosinska2018beyond,wei2017road} proposed topology-related losses to take topology into consideration.

Recent studies of road extraction focus on exploiting more features. For example, 
\cite{gao2018end} designed a multiple feature pyramid network with dilated convolutional layers to enlarge the receptive field. 
Some studies~\cite{demir2018deepglobe,qiangqiang2020automatic,zhou2018d,zhou2020bt} took advantages of Atrous Spatial Pyramid Pooling~\cite{chen2017deeplab} to reduce the loss of spatial information and enhance context awareness. 
In particular, D-LinkNet~\cite{zhou2018d} won the championship of DeepGlobe 2018~\cite{demir2018deepglobe}. Some studies~\cite{jegou2017one,li2020topology,ren2020capsunet,xu2018road} 
employed attention modules to capture long-range dependencies and obtain better features,
while some other studies focused on the road extraction task with the support of other types of data, e.g. satellite images and 
GPS data~\cite{sun2019leveraging,wu2020deepdualmapper}, 
satellite and lidar data~\cite{parajuli2018fusion} 
or pure GPS data~\cite{ruan2020learning}.
Different from these these existing studies, our work targets a new yet practical setting for road extraction, i.e. we make use of partial road maps for the task given that they are of high quality and publicly available.

\subsection{Data Fusion}
\noindent For tasks where the satellite images and some other data types such as lidar data, laser data, and GPS data are involved, data fusion is normally adopted. 
There are mainly three types of techniques for data fusion, namely inputs fusion, outputs fusion and features fusion. 
Some studies~\cite{nahhas2018deep,sun2019leveraging,sun2018combining} 
concatenated satellite images and other data types of image such as aerial lidar images as inputs for inputs fusion.
Outputs fusion~\cite{gadiraju2020multimodal} can be achieved either using extra layers to refine predictions from different
sources~\cite{gadiraju2020multimodal,rudner2019multi3net} 
or directly taking an average of the predictions to obtain final outputs. Features fusion fuses feature maps from different branches. 
For example, 
\cite{cao2018integrating} embedded features from one branch to another, while 
studies~\cite{benedetti2018m,li2020collaborative,parajuli2018fusion,wu2020deepdualmapper}
utilized an extra branch to process the fused features. 
We design a features fusion approach in this paper. Different from the existing features fusion methods on satellite images, we take the branch for satellite images as an auxiliary one and that for partial road maps as the main one and fuse the features from the satellite images to the main branch. 

\subsection{Road Inpainting}
\noindent As introduced by \cite{chen2019blind}, road inpainting is a task of inpainting a road map (e.g. denoising and fixing broken road segments). \cite{batra2019improved} used road inpainting as a post-processing step for refining extracted road maps. 
\cite{zhang2019road} made use of satellite images for road inpainting, which concatenates satellite images and some noisy/broken road maps as inputs. We note that the noisy/imperfect road maps that are taken as inputs for the road inpainting task are by nature different from the partial road maps used in this work. For example, the former usually does not have some entire road segments missing and thus the task is easier and feasible even without other data sources such as satellite images taken as inputs, but the latter may involve the majority of road segments missing and thus it is not feasible to generate complete road maps solely from the partial ones in general.

\section{Methodology}
\label{sec:method}

\noindent 
We first present the framework, then introduce the newly-designed Gated Self-Attention Module (GSAM), which could capture long-range semantics and control the information flow from the two data sources in a complementary manner, and finally introduce the employed loss functions including the new Missing Part (MP) loss.

\subsection{Model Architecture}

\noindent The framework of P2CNet is presented in Figure \ref{p2cnet}. P2CNet has two branches, one takes the satellite images as inputs and is called the \emph{satellite branch} (blue) and the other takes the partial road maps as inputs and is called the \emph{partial branch} (orange).
Each branch involves an encoder and a decoder, and for both of them we employ DeepLabV3+~\cite{chen2018encoder} with the backbone of ResNet-34 as the basic model due to its small number of parameters and good performance as verified in our experiments. 

In particular, we develop a module for better utilizing and fusing the features outputted by the encoders, which is referred as the Gated Self-Attention Module (GSAM).

First, inspired by Dual Self-Attention~\cite{fu2019dual}, we decide to apply two channel-wise self-attentions to capture global context over features for both satellite and partial branches. Then, we fuse their affinity matrices to provide a mutual guidance and emphasize interdependent feature maps. Different from ~\cite{fu2019dual} that applies a self-attention mechanism on both spatial ($H$, $W$) and channel ($C$) dimensions, we only consider channel dimension in this work. There are two reasons that we employ channel-wise but not spatial attentions in GSAM. First, channel-wise attentions could save much memory, as the shape of its affinity matrix is $C \times C$, while that of spatial attentions is $HW \times HW$. Second, partial road maps usually involve a small number of road pixels, and many extracted feature maps would be close to all-zero matrices, making it less useful to apply spatial attentions. With channel-wise self-attentions, various semantics could be highlighted per channel to improve the feature representation of specific semantics.

Second, a gate mechanism is adopted to further control the information flow in a complementary manner.
We then take the generated road map from the partial branch as the output, i.e. we take the partial branch as the main branch. We note that the task of generating complete road maps from partial road maps without features from satellite images is not feasible (as verified via experiments) and needs to be guided with features extracted from the satellite images.%

\subsection{Gated Self-Attention Module (GSAM)}

\noindent Self-attention mechanism has shown its advantages in capturing long-range dependencies and improving the feature representation of specific semantics~\cite{fu2019dual}, while gate mechanism is widely adopted in multimodal data fusion to fuse different features. We build the GSAM module to capture rich features from satellite images and partial road maps with self-attentions and then fuse them in a complementary manner with a gate. 
Note that there are also 4 GSAMs set right behind the last 4 stages of ResNet-34 in the partial branch's encoder.
GSAM is presented in Figure \ref{gsam}, and we formally describe it as follows.

\subsubsection{Channel-wise Self-Attention}

\noindent It takes feature maps $F_S \in \mathbb{R}^{C \times H \times W}$ and $F_P \in \mathbb{R}^{C \times H \times W}$ from the satellite branch and the partial branch, respectively, as inputs. The affinity matrix $A_S \in \mathbb{R}^{C \times C}$ and $A_P \in \mathbb{R}^{C \times C}$ are derived directly from the inputs by reshaping $F_S$ and $F_P$ both to $\mathbb{R}^{C \times HW}$, then performing a matrix multiplication between them and their transpose, and finally applying a softmax operation:
\begin{equation}
    a_X^{ji} = \frac{e^{F_X^i \cdot F_X^j}}{\sum_{i=1}^{C}e^{F_X^i \cdot F_X^j}}
\end{equation}
where $X \in \{S, P\}$, $a_X^{ji}$ is the element with coordinate of $(j, i)$ in $A_X$, measuring the impact of the $i^{th}$ channel on the $j^{th}$ channel. Then, to achieve the goal of mutual guidance, an element-wise multiplication on $A_S$ and $A_P$ is performed to obtain the fused affinity matrix $A_{fuse} \in \mathbb{R}^{C \times C}$. After that, two matrix multiplications between the transpose of $A_{fuse}$ and $F_S, F_P$ are performed, and their results are reshaped to $\mathbb{R}^{C \times H \times W}$. The enhanced features $F_S^{\prime} \in \mathbb{R}^{C \times H \times W}$ and $F_P^{\prime} \in \mathbb{R}^{C \times H \times W}$ are calculated as:
\begin{align}
A_{fuse} &= A_S \odot A_P \\
F_X^{\prime j} &= \beta_X \sum_{i=1}^{C}(a_{fuse}^{ji}F_{X}^{i}) + F_{X}^{j}
\end{align}
where $j$ means each position of feature maps, $\beta_{X}$ represents two learnable weights initialized as 0 for $S$ and $P$, $a_{fuse}^{ji}$ is the element with the coordinate of $(j, i)$ in $A_{fuse}$.

\subsubsection{Gate Mechanism}

\noindent Then, $F_S^{'}$ and $F_P^{\prime}$ are concatenated in the channel dimension to be the input $F_C \in \mathbb{R}^{2C \times H \times W}$ of convolution blocks. The convolution block consists of a convolution layer, a batch normalization layer, followed by a ReLU activation. The kernel sizes of the convolution layers are all $3 \times 3$, and their output channels are all $C$. After convolution, two feature maps $F_{C_S} \in \mathbb{R}^{C \times H \times W}$ and $F_{C_P} \in \mathbb{R}^{C \times H \times W}$ are generated. Next, $F_{C_S}$ and $F_{C_P}$ are unsqueezed to have a new dimension, and concatenated along it. A softmax layer is further attached to offer complementary fusion and result in $F_C^{\prime} \in \mathbb{R}^{2 \times C \times H \times W}$. Then, two element-wise multiplications are used to obtain the final features $F_S^{\prime \prime} \in \mathbb{R}^{C \times H \times W}$ and $F_P^{\prime \prime} \in \mathbb{R}^{C \times H \times W}$. The fused features $F_{fuse} \in \mathbb{R}^{C \times H \times W}$ are the sums of $F_S^{\prime \prime}$ and $F_P^{\prime \prime}$:
\begin{equation}
    F_C^{\prime}[0, :, :, :] + F_C^{\prime}[1, :, :, :] = 1
\end{equation}
\begin{equation}
    F_{fuse} = \underbrace{F_S^{\prime} \odot F_C^{\prime}[0, :, :, :]}_{F_S^{\prime \prime}} + \underbrace{F_P^{\prime} \odot F_C^{\prime}[1, :, :, :]}_{F_P^{\prime \prime}}
\end{equation}
where $F_C^{\prime}$ is the output of the convolution blocks, $[0, :, :, :]$ and $[1, :, :, :]$ mean the $1^{st}$ and $2^{nd}$ channels. Finally, $F_{fuse}$ is forwarded back to the partial branch for further processing.

\begin{figure}[!t]
\centering
\includegraphics[width=1\linewidth]{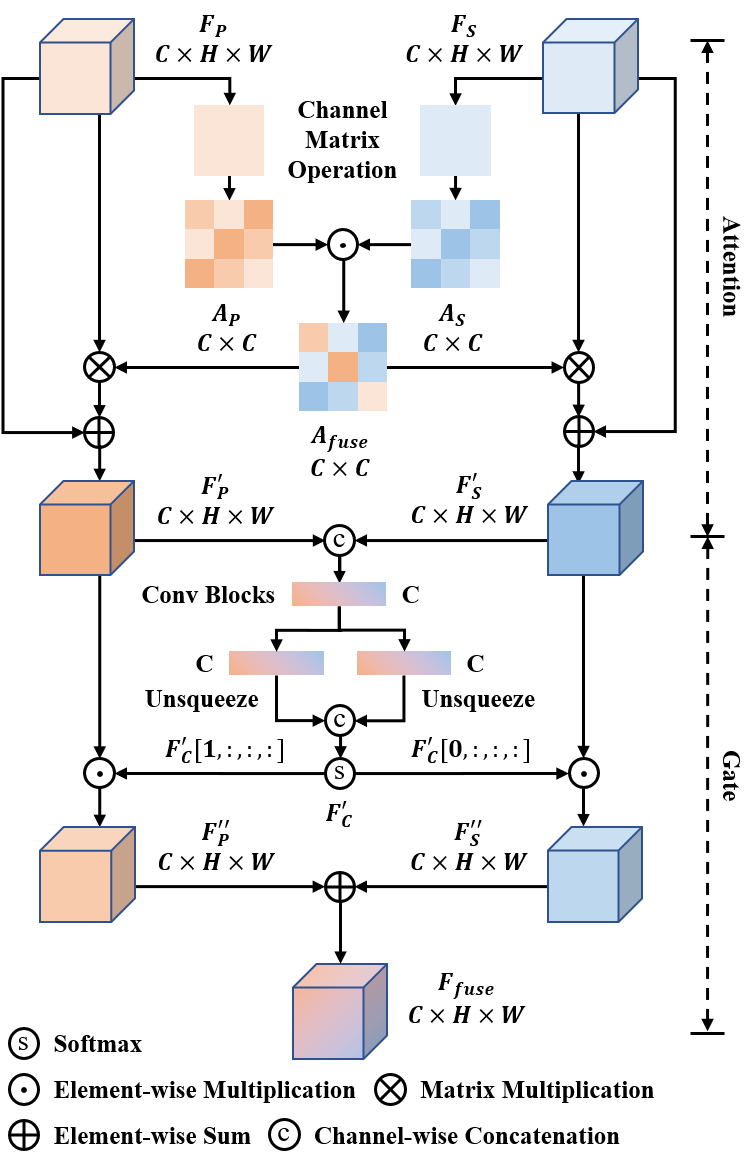}
\caption{Details of GSAM.}
\label{gsam}
\end{figure}

\subsection{Loss Functions}

\noindent We employ a combination of binary cross entropy (BCE) loss and Dice loss as the basic loss functions of our model. BCE loss is widely accepted in binary semantic segmentation tasks, while Dice loss is usually used when the classes are imbalanced.
In the road extraction task, the number of road pixels is far less than that of background pixels. These loss functions are defined as follows:
\begin{align}
\mathcal{L}_X^{BCE} &= - \frac{1}{N}\sum_{i=1}^{N}(I_C^i \log M_X^i + (1 - I_C^i)(1 - \log M_X^i)) \\
\mathcal{L}_X^{Dice} &= 1 - \frac{2|M_X \cap I_C|}{|M_X| + |I_C|}
\end{align}
where $X \in \{S, P\}$, $N = H \times W$, $I_C \in \mathbb{R}^{1 \times H \times W}$ is the complete road map, $M_S \in \mathbb{R}^{1 \times H \times W}$, and $M_P \in \mathbb{R}^{1 \times H \times W}$ are the generated road maps from satellite and partial branches, respectively.

In addition, we propose a new loss called the \emph{missing part} (MP) loss to further exploit the partial road maps. The MP loss is defined based on those missing road pixels from the partial road maps as follows:
\begin{align}
I_{MP} &= I_C - I_P \\
\mathcal{L}_S^{MP} &= - \frac{1}{|\Omega_{MP}|}\sum_{i \in \Omega_{MP}}(I_{MP}^i \log M_S^i)
\end{align}
where $I_{MP}$ is the missing part of the road map, $I_P$ is the partial road map, $\Omega_{MP}$ is the road pixels set of $I_{MP}$, and $|\cdot|$ means the number of elements in the set.
The rationale of MP loss is to assign more weights to the missing road pixels when performing back propagation. Note that the MP loss corresponds to a BCE loss on the missing road pixels. We do not define the MP loss as a Dice loss on the missing road pixels since the latter is sensitive to small regions and makes the training process unstable~\cite{zhang2021rethinking}. For the complete road map, which has $1024^2$ pixels, the Dice loss works well, but for the missing part, which occupies a very small number of pixels only, e.g. several hundreds or fewer pixels, a Dice loss would not work well.

Finally, the loss of satellite branch $\mathcal{L}_{S}$, the loss of partial branch $\mathcal{L}_{P}$ and the total loss $\mathcal{L}_{total}$ are as follows:
\begin{equation}
    \mathcal{L}_{total} = \underbrace{\mathcal{L}_S^{BCE} + \mathcal{L}_S^{Dice} + \lambda\mathcal{L}_S^{MP}}_{\mathcal{L}_S} + \underbrace{\mathcal{L}_P^{BCE} + \mathcal{L}_P^{Dice}}_{\mathcal{L}_P}
    \label{equ:MP-loss}
\end{equation}
where $\lambda$ is a hyperparameter.
Note that we incorporate the MP loss in the satellite branch only, which we explain as follows. 
The MP loss would help to put more weights on those missing road pixels and thus extract them out correctly, but at the same time, it would lower down the weights on other pixels including those near the roads in the partial road maps, resulting in some potential misclassifications. Specifically, for those pixels near to roads, they tend to be classified to be road pixels (since they correspond to roads in the satellite image) but their corresponding labels are background pixels (since only those pixels on the centerlines of roads are labeled as road pixels while others as background pixels). By incorporating the MP loss in the satellite branch only, we achieve (1) some additional weights have been put on those missing road pixels as we aim to and (2) the misclassifications are avoided since they can only happen in the satellite branch and the generated road map in the partial branch is taken as the output.
We conduct detailed experiments on different strategies of using the MP loss strategies in Section \ref{ablation_sec}, which verify the superiority of the strategy as implemented by Equation~(\ref{equ:MP-loss}). 

\section{Experiments}
\label{sec:experiments}

\subsection{Dataset}

\noindent By following the existing studies~\cite{batra2019improved,tran2020pp,van2020city} and ~\cite{bastani2018roadtracer}, we use the SpaceNet\footnote{\url{https://spacenet.ai/spacenet-roads-dataset/}}~\cite{van2018spacenet} and OSM\footnote{\url{https://github.com/mitroadmaps/roadtracer/}} datasets. SpaceNet contains 1300px$\times$1300px satellite images and their corresponding complete road maps, which cover four cities, including Las Vegas, Paris, Shanghai and Khartoum. 
After removing those satellite images, for which there are no complete road maps, we have 283, 257, 1,028, and 981 satellite images for Khartoum, Paris, Shanghai and Las Vegas, respectively. In total, we have 2,549 satellite images and convert them to 8-bit RGB images. And the OSM dataset contains 1024px$\times$1024px satellite images of 40 cities from Google Maps and the road maps from OSM.

We define a parameter called \emph{completeness ratio} $\alpha$ of a partial road map, which represents the ratio of the number of road pixels in the partial road map and that in the complete road map. To obtain the partial road maps, we randomly erase complete road segments until the remaining road pixels are below $\alpha$.

The preprocessing step is uniform, and we only take SpaceNet as an example. For partial road maps, we choose to generate them from the complete ones by erasing some roads, but not to use the road maps from OSM which are incomplete for two reasons. First, the road maps from SpaceNet and those from OSM are from two different sources and extra efforts are needed to align them to each other, which could be tedious. Second, with the road maps from OSM used as partial road maps, we cannot control the extent to which the roads are missing for studying the effects of the missing roads on the performance of the proposed models. 

\begin{figure}[!t]
\centering
\includegraphics[width=0.8\linewidth]{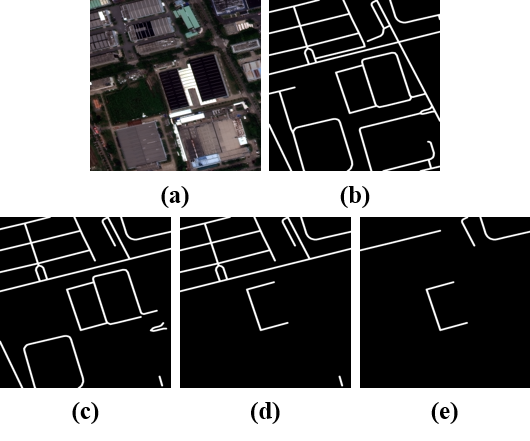}
  \caption{A sample of the datasets. (a) Satellite image; (b) Complete road map; (c) 75\% partial road map; (d) 50\% partial road map; (e) 25\% partial road map.}
\label{data_sample}
\end{figure}

To generate a partial road map with the completeness ratio $\alpha$, we iteratively drop a road from the road map until the ratio of road pixels in the remaining road map and that in the original one becomes below $\alpha$. Note that we do not define the completeness ratio as the ratio of the numbers of roads since roads can have different lengths and a partial road map with one long road missing would be significant different from that with one short road missing. We prepare five sets of partial road maps as follows: for the first four sets, we generate them by setting $\alpha$ as 0\%, 25\%, 50\%, and 75\% and for the fifth set, we generate each partial road map by randomly sampling it from one of the first four sets except for the 0\% set and call the set \emph{mix}. 
An illustration of some partial road maps is shown in Figure~\ref{data_sample}.

\subsection{Training Details}

\noindent We randomly split the dataset into train, validation and test sets, with ratios of 0.8, 0.1 and 0.1, respectively. We use random crops with size of 1024$\times$1024 for the train set and the crop with size of 1024$\times$1024 in the center for validation and test. Some data augmentation methods, e.g. random horizontal flip, shift, Gaussian noises, and brightness are applied to the train set. The proposed model is implemented using PyTorch and adopts the Adam optimizer with an initial learning rate of $2 \times 10^{-4}$ and a momentum of 0.5 for optimization.
The training batch size is set as 4.
After training each epoch, validation is performed, and a learning rate scheduler is further utilized based on this. If the $mIoU$ score does not increase continuously for 5 epochs, the learning rate would be multiplied by 0.2. In addition, the training process would be terminated if the $mIoU$ score does not improve for 20 consecutive epochs.
Then, we use the model, whose validation $mIoU$ score is the best, to perform testing.

\subsection{Baselines}

\noindent We select 10 baselines including
PSPNet~\cite{zhao2017pyramid}, PAN~\cite{li2018pyramid}, FPN~\cite{kirillov2017unified}, CasNet~\cite{cheng2017automatic}, LinkNet~\cite{chaurasia2017linknet}, DLinkNet~\cite{zhou2018d}, DeepLabV3~\cite{chen2017rethinking}, DeepLabV3+~\cite{chen2018encoder}, RoadCon~\cite{batra2019improved}, and SGCN~\cite{zhou2021split}, for comparisons. 
In addition, we consider McGAN~\cite{zhang2019road}, which was originally designed for road inpainting, for comparison, for which we use partial road maps as noisy ones. 
For baseline methods, we adapt them to the proposed setting with inputs fusion. All methods use the same inputs (namely partial maps and satellite images) for fair comparison.
Moreover, we take ResNet-34, which is pretrained on ImageNet, as the backbone for all methods, and each model is trained and tested with 1024$\times$1024 image patches.

\begin{table*}[!t]
\caption{Comparisons of models for road extraction. All models are trained and tested on the mix SpaceNet dataset; The table is sorted based on $mIoU$ in ascending order. (sat)/(par) means outputs from satellite/partial branch.}
\label{performance_comparisons_spacenet}
\centering
\resizebox{.6\linewidth}{!}{
    \begin{tabular}{lrrrrrr}
    \hline
    \multirow{2}{*}{Method} & \multicolumn{6}{c}{SpaceNet} \\
    & $P$ & $R$ & $F1$ & $APLS$ & $IoU$ & $mIoU$\\
    \hline
    CasNet~\cite{cheng2017automatic} & 0.8056 & 0.7653 & 0.7849 & 0.6105 & 0.6460 & 0.8094 \\
    PSPNet~\cite{zhao2017pyramid} & 0.8275 & 0.7679 & 0.7966 & 0.6597 & 0.6619 & 0.8183 \\
    RoadCon~\cite{batra2019improved} & 0.8419 & 0.7758 & 0.8075 & 0.7040 & 0.6772 & 0.8266 \\
    McGAN~\cite{zhang2019road} & 0.8388 & 0.7990 & 0.8178 & 0.7420 & 0.6917 & 0.8345 \\
    FPN~\cite{kirillov2017unified} & 0.8332 & 0.8075 & 0.8202 & 0.7297 & 0.6952 & 0.8361 \\
    DeepLabV3~\cite{chen2017rethinking} & 0.8371 & 0.8052 & 0.8208 & 0.7486 & 0.6961 & 0.8367 \\
    PAN~\cite{li2018pyramid} & 0.8365 & 0.8082 & 0.8221 & 0.7078 & 0.6979 & 0.8376 \\
    DeepLabV3+~\cite{chen2018encoder} & 0.8378 & 0.8085 & 0.8229 & 0.7359 & 0.6990 & 0.8382 \\
    SGCN~\cite{zhou2021split} & 0.8377 & 0.8100 & 0.8237 & 0.7415 & 0.7002 & 0.8388 \\
    LinkNet~\cite{chaurasia2017linknet} & 0.8412 & 0.8075 & 0.8240 & 0.7384 & 0.7007 & 0.8392 \\
    DLinkNet~\cite{zhou2018d} & 0.8385 & 0.8116 & 0.8248 & 0.7445 & 0.7019 & 0.8398 \\
    \hline
    P2CNet & \bf 0.8424 & \bf 0.8150 & \bf 0.8285 & \bf 0.7489 & \bf 0.7071 & \bf 0.8426 \\
    \hline
    \end{tabular}
}
\end{table*}

\begin{table*}[!t]
\caption{Comparisons of models for road extraction. All models are trained and tested on the mix OSM dataset; The table is sorted based on $mIoU$ in ascending order. (sat)/(par) means outputs from satellite/partial branch.}
\label{performance_comparisons_osm}
\centering
\resizebox{.6\linewidth}{!}{
    \begin{tabular}{lrrrrrr}
    \hline
    \multirow{2}{*}{Method} & \multicolumn{6}{c}{OSM} \\
    & $P$ & $R$ & $F1$ & $APLS$ & $IoU$ & $mIoU$\\
    \hline
    PSPNet~\cite{zhao2017pyramid} & 0.8365 & 0.8044 & 0.8201 & 0.7234 & 0.6851 & 0.8311 \\
    PAN~\cite{li2018pyramid} & 0.8579 & 0.7851 & 0.8199 & 0.6449 & 0.6947 & 0.8313 \\
    DeepLabV3~\cite{chen2017rethinking} & 0.8361 & 0.8054 & 0.8205 & 0.7368 & 0.6956 & 0.8314 \\
    CasNet~\cite{cheng2017automatic} & 0.8701 & 0.8030 & 0.8331 & 0.6871 & 0.7140 & 0.8424 \\
    McGAN~\cite{zhang2019road} & 0.8815 & 0.8013 & 0.8395 & 0.7843 & 0.7234 & 0.8474 \\
    FPN~\cite{kirillov2017unified} & 0.8655 & 0.8192 & 0.8417 & 0.7625 & 0.7267 & 0.8490 \\
    RoadCon~\cite{batra2019improved} & 0.8762 & 0.8167 & 0.8454 & 0.7779 & 0.7322 & 0.8522 \\
    DeepLabV3+~\cite{chen2018encoder} & 0.8662 & 0.8353 & 0.8505 & 0.7821 & 0.7398 & 0.8562 \\
    DLinkNet~\cite{zhou2018d} & 0.8772 & 0.8258 & 0.8507 & 0.7793 & 0.7402 & 0.8566 \\
    SGCN~\cite{zhou2021split} & 0.8760 & 0.8270 & 0.8508 & 0.7803 & 0.7403 & 0.8566 \\
    LinkNet~\cite{chaurasia2017linknet} & \bf 0.8835 & 0.8233 & 0.8523 & 0.7801 & 0.7427 & 0.8580 \\
    \hline
    P2CNet & 0.8788 & \bf 0.8430 & \bf 0.8605 & \bf 0.8028 & \bf 0.7552 & \bf 0.8648 \\
    \hline
    \end{tabular}
}
\end{table*}

\subsection{Evaluation Metrics}

\noindent Six metrics are applied in the experiments: precision $P$, recall $R$, F1 score $F1$, APLS score $APLS$~\cite{van2018spacenet}, intersection over union $IoU$ and mean intersection over union $mIoU$. $P$ measures the ratio of correctly extracted road pixels among all extracted road pixels, $R$ measures the ratio of correctly extracted road pixels among all actual road pixels, and $F1$ is a harmonic mean of $P$ and $R$. 
$APLS$ is a topology-oriented metric and measures the differences of optimal path lengths between nodes in generated and ground truth road graphs. $IoU$ calculates the ratio of the overlapping area to the union area of the actual and extracted road pixels and $mIoU$ is the average of $IoU$'s for road pixels and background pixels.

Road extraction is a binary segmentation task, where only 2 classes involve, i.e., foreground and background. All metrics except $APLS$ could be used as segmentation-related metrics, and could be calculated via confusion matrix, which are shown as follow.
\begin{align}
P &= \frac{TP}{TP+FP} \\
F &= \frac{TP}{TP+FN} \\
F1 &= \frac{2TP}{2TP+FP+FN} \\
IoU &= IoU_f = \frac{TP}{TP+FP+FN} \\
IoU_b &= \frac{FN}{FN+TP+TN} \\
mIoU &= \frac{1}{2}(IoU_f+IoU_b)
\end{align}
where $TP$, $TN$, $FP$, $FN$ represent true positive, true negative, false positive, and false negative, respectively. $IoU_f$ represents the $IoU$ of road, while $IoU_b$ is that of background.

Particularly, for $APLS$, it does not treat the extracted road maps as an image. Instead, it extracts the image's skeleton, and builds a graph based on the skeleton. Then, it measures the sum-up differences in the lengths of the shortest paths between graph nodes of the newly-built graph, and the exact ground truth. In this way, $APLS$ could measure topology difference between the two graphs well. For example, if there exists a road segment between two points in the road map, and the road segment is broken in the extracted graph, then we might still be able to travel between two points, but the length of the shortest path would become much larger.

\section{Results}
\label{sec:results}

\subsection{Performance Comparisons}

\subsubsection{Quantitative Results}

\noindent The results are shown in Table~\ref{performance_comparisons_spacenet} and Table~\ref{performance_comparisons_osm}.

For the selected models, we set the number of input channels to 4 and concatenate satellite images and partial road maps along channel dimension as inputs, as inspired by McGAN. Besides, inputs fusion could obtain quite good results, according to later experiments on the ways of fusion.
In particular, for McGAN, we use a pretrained DeepLabV3+ model as the generator.
The results show that its discriminators do not work well in our setting, which cause a slight performance decrease of 1.04\% on $mIoU$, compared with DeepLabV3+. 
According to the results, our P2CNet model
achieves the best results consistently across 6 metrics (including the topology-oriented one APLS), on 2 datasets from different sources and of different qualities (OSM and SpaceNet), and among 11 methods, e.g., for $IoU$ score, P2CNet could outperform the best baseline, say, LinkNet, by 1.27\% on the OSM mix dataset. Besides, the $APLS$ score of P2CNet is 1.85\% better than the second.

In addition, in our P2CNet, the $IoU$ scores of the predictions based on the partial branch are 18.60\% and 23.05\% higher than those based on the satellite branch in SpaceNet and OSM, respectively. Such huge gaps could show the great importance of partial road maps that could help to reduce the difficulty of road extraction greatly.

\subsubsection{Model Complexity}

\begin{table}[!t]
\caption{Comparisons of Model Complexity.}
\label{complexity}
\centering
\resizebox{.9\linewidth}{!}{
    \begin{tabular}{lrrrrrr}
    \hline
    Method & \#Branch & \#Parameters (M) & MACs (G) \\
    \hline
    DLinkNet~\cite{zhou2018d} & 1 & 31.21 & 4.68 \\
    DeepLabV3+~\cite{chen2018encoder} & 1 & 22.44 & 6.10 \\
    SGCN~\cite{zhou2021split} & 1 & 32.78 & 26.04 \\
    \hline
    P2CNet & 2 & 61.04 & 28.04\\
    \hline
    \end{tabular}
}
\end{table}

\begin{figure*}[!t]
\centering
\includegraphics[width=1\linewidth]{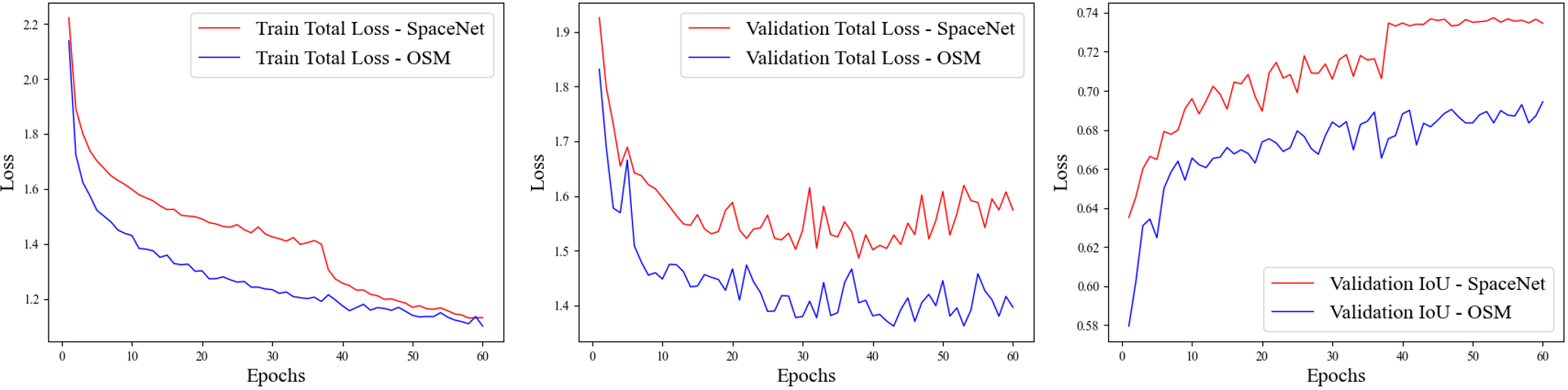}
\caption{Loss and IoU curves of P2CNet on mix SpaceNet and OSM datasets.}
\label{convergence}
\end{figure*}

As shown in Table~\ref{complexity}, we use {\em thop}\footnote{\url{https://github.com/Lyken17/pytorch-OpCounter}} to calculate the number of parameters and MACs (Multiply–ACcumulate operations). Both satellite images and partial road maps are taken as the inputs, and their spatial resolutions are uniformly set as 224$\times$224. 
We have the following observations.
(1) Our P2CNet consists of two branches. The satellite branch is exactly a DeepLabV3+ model, and the partial branch is built upon DeepLabV3+ (with novel attention-based GSAM modules). Hence, 
our model has more parameters and MACs than DeepLabV3+.
(2) The number of MACs of our P2CNet (two-branch) is larger than that of SGCN (one-branch) by 2G only, which shows the effectiveness of our design.

\subsubsection{Model Convergence}

To show the convergence speed of our P2CNet, we show the results of the total training and validation losses, as well as the validation IoU scores, on both mix SpaceNet and OSM datasets in Figure~\ref{convergence}.
We could observe that P2CNet could converge fast, e.g., P2CNet could converge within 40 epochs on both datasets.

\subsubsection{Qualitative Results}

\begin{figure*}[!t]
\centering
\includegraphics[width=1\linewidth]{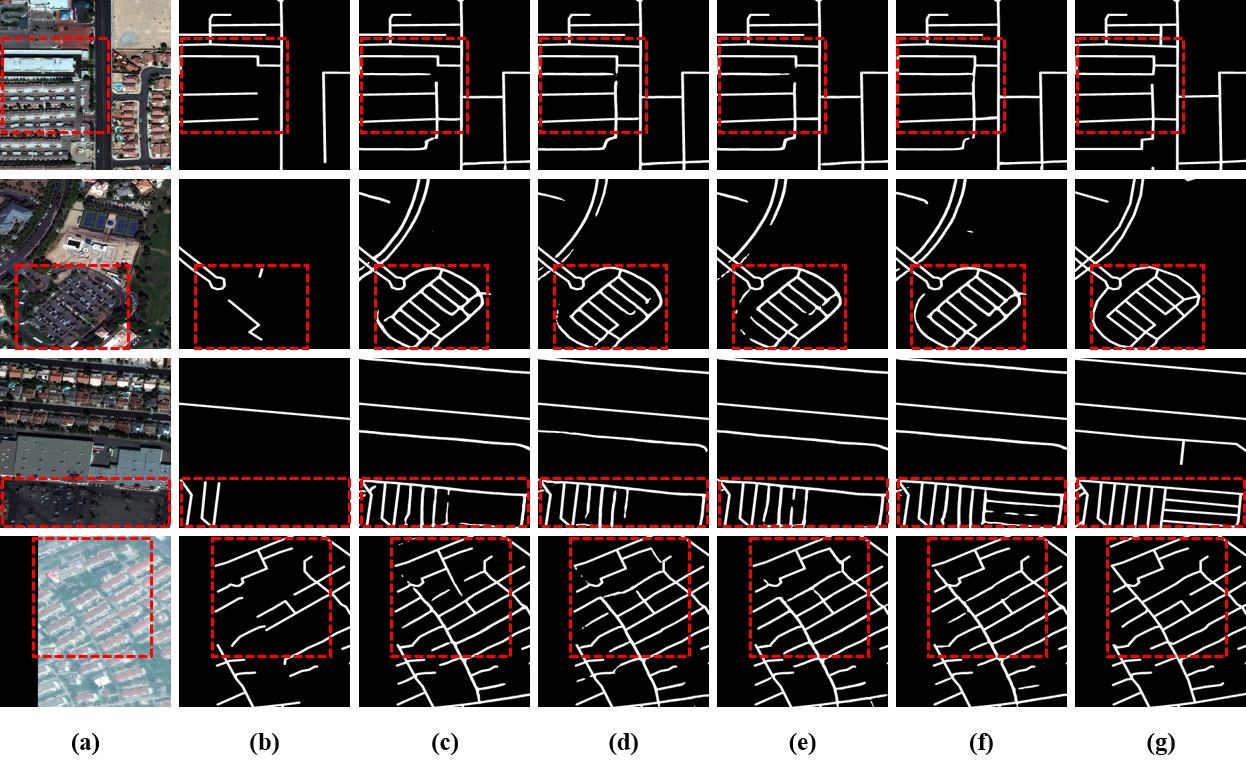}
\caption{Qualitative results of mix SpaceNet. (a) Satellite image; (b) Partial road map (SpaceNet); (c) DeepLabV3+; (d) LinkNet; (e) DLinkNet; (f) P2CNet (ours); (g) Complete road map (SpaceNet).}
\label{visualization}
\end{figure*}

\begin{figure*}[!t]
\centering
\includegraphics[width=1\linewidth]{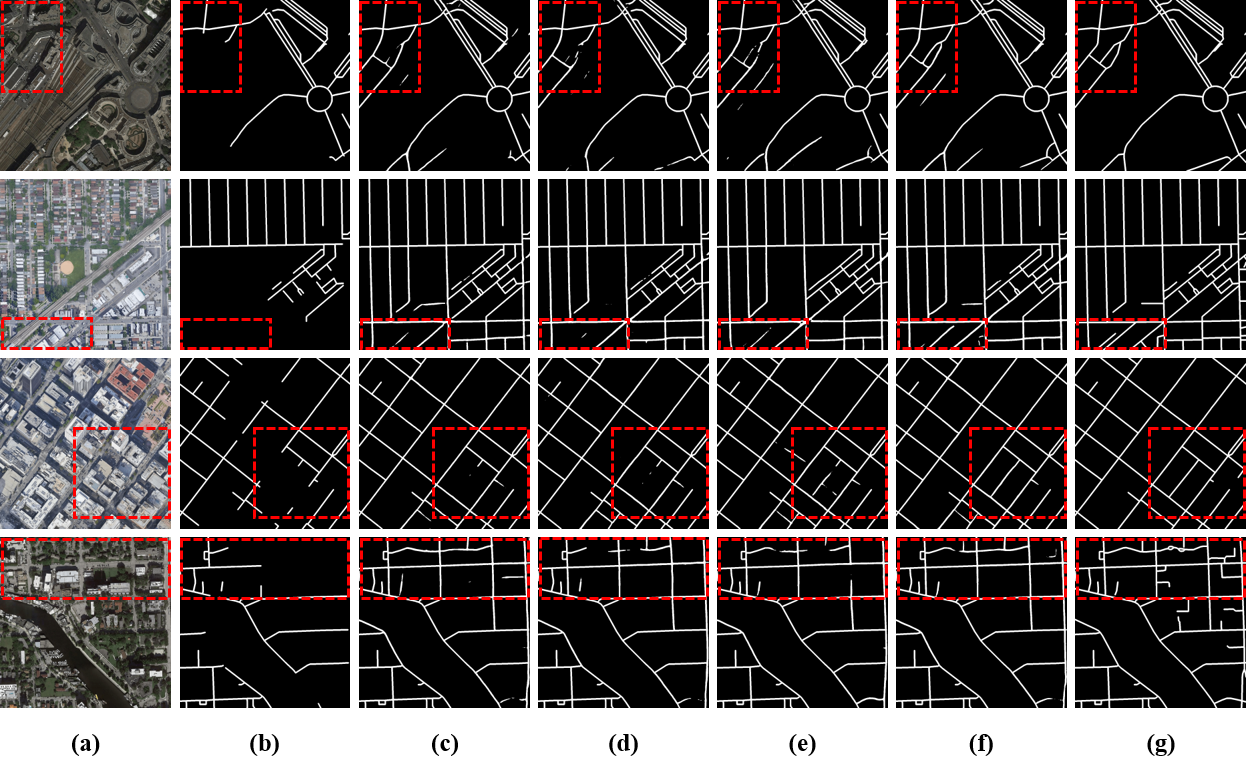}
\caption{Qualitative results of mix OSM. (a) Satellite image; (b) Partial road map (OSM); (c) DeepLabV3+; (d) LinkNet; (e) DLinkNet; (f) P2CNet (ours); (g) Complete road map (OSM).}
\label{visualization_osm}
\end{figure*}

\noindent We visualize four sets of satellite images, partial road maps, complete road maps and the generated road maps by different models from SpaceNet in Figure~\ref{visualization}, and OSM in Figure~\ref{visualization_osm}. We use red rectangles to highlight some areas that are difficult to be detected, and it could be observed that our P2CNet works well on these areas. This is due to the effectiveness of our proposed modules, say, GSAM and MP loss. With the help of channel-wise self-attention and the gate mechanism, the proposed GSAM could filter out the useless and sparse features in partial road maps, and the remaining useful road features could be better referred by the module, and then build up stronger relationship with the features extracted from the satellite images.

An explanation for this is as follow, take the third row of Figure~\ref{visualization} as an example, the area inside the red rectangle are quite hard to be classified as roads. We could observe that in column (b), there are some existing roads in the left side of rectangle, and these existing parts have perceptually similar features as the other parts inside the rectangle, according to column (a). Then, the existing of partial road maps appear to act as clues, and could encourage the model to analyze the relationships between these areas and other areas, say, the right part of the rectangle has similar features as parts with partial road maps, hence, these areas would probably contain roads.

\subsection{Ablation Study} \label{ablation_sec}

\noindent To study the P2CNet model in detail, extensive experiments are conducted in this section, including different ways of fusion, MP loss strategies, ablation study on modules, and robustness of the proposed model.
\subsubsection{Fusion Strategies}

\noindent We investigate four different fusion strategies based on DeepLabV3+ with the backbone of ResNet-34: {\bf (1) Outputs Augmentation} performs fusion directly by taking the union of binarized outputs and partial road maps as the final outputs; {\bf (2) Inputs Fusion} feeds the model with the concatenations of satellite images and partial road maps; {\bf (3) Outputs Fusion} is similar to Outputs Augmentation, but instead of using Sigmoid activation at last, it uses two 3$\times$3 convolution layers to convert the inactivated outputs and partial road maps into the same space, concatenates the outputs of convolution layers, and uses another 1$\times$1 convolution layer to obtain the final outputs, followed by a Sigmoid activation; {\bf (4) Features Fusion} is our P2CNet with all GSAM replaced with 3$\times$3 convolution layers. The model of Features Fusion is two-branch, while others are one-branch.
More details of these four fusion strategies can be found in the appendix.

\begin{table}[!t]
\caption{Results of fusion strategies. All models are trained on the mix dataset; $\alpha_{test}$ means $\alpha$ adopted for testing. (sat)/(par) in Feature Fusion means outputs from satellite/partial branch. {\bf Bold} scores are the best, while {\underline{scores}} with underline are the second best.}
\label{fusion_strategy}
\centering
\begin{tabular}{lrrr}
\hline
Method & $\alpha_{test}$ & $F1$ & $IoU$ \\
\hline
Outputs Augmentation & mix & 0.7711 & 0.6274 \\
Outputs Fusion & mix & 0.7782 & 0.6369 \\
Inputs Fusion & mix & \underline{0.8229} & \underline{0.6990} \\
Features Fusion (sat) & mix & 0.6843 & 0.5201 \\
Features Fusion (par) & mix & \bf 0.8248 & \bf 0.7019 \\
\hline
Outputs Augmentation & 0\% & \bf 0.6909 & \bf 0.5277 \\
Outputs Fusion & 0\% & 0.6597 & 0.4922 \\
Inputs Fusion & 0\% & 0.6556 & 0.4876 \\
Features Fusion (sat) & 0\% & \underline{0.6843} & \underline{0.5201} \\
Features Fusion (par) & 0\% & 0.6688 & 0.5024 \\
\hline
\end{tabular}
\end{table}

We uniformly train these models on SpaceNet mix dataset, and then test them on both mix and 0\% datasets. The results are shown in Table \ref{fusion_strategy}.
We observe that Outputs Augmentation outperforms others on the 0\% dataset, but does not performs well on mix dataset, because the partial road maps are not involved in training process in this case. In addition, Inputs Fusion does not have a good generalization ability, and the reason is that partial road maps are binary images and if they are directly used as parts of the inputs, as Inputs Fusion does, the model would focus on how to reserve such information instead of inferring the road maps from satellite features.
We employ Features Fusion since it achieves the best performance when considering both mix and 0\% datasets, i.e. it generalizes well on both cases,
because individual loss is calculated and back-propagated on the satellite branch, so the features of satellite images could be well captured, and the influence of partial road maps on satellite images would not be as severe as that in Inputs Fusion.

\subsubsection{Loss Strategies}

\noindent The two-branch Features Fusion model is used as the basic model. There are three possible ways to employ the MP loss: {\bf (1) Both}, add the MP loss to both the satellite and partial branches; {\bf (2) Par}, only add the loss to the partial branch; {\bf (3) Sat}, only add to the satellite branch. As the missing parts of a road map only occupy a very small portion of pixels, small changes of the weight $\lambda$, e.g. from 0.9 to 1, might not affect the performance much and the influence of randomness caused by data augmentation might blow up, we only try weights with relatively large intervals in this paper. To amplify the impact of the MP loss on metrics, we use $\lambda = $ 1, 10, 20, 30 to demonstrate the differences among the three loss strategies.

\begin{table}[!t]
\caption{Experiments on MP loss strategies. All models are trained and tested on mix dataset (SpaceNet).}
\label{mp_loss_strategy}
\centering
\begin{tabular}{lrrrr}
\hline
Method & $\lambda$ & $R$ & $F1$ & $IoU$ \\
\hline
Both & 1 & 0.8241 & 0.8236 & 0.7001 \\
& 10 & 0.8747 & 0.8057 & 0.6746 \\
& 20 & 0.8978 & 0.7594 & 0.6121 \\
& 30 & \bf 0.9179 & 0.7353 & 0.5814 \\
\hline
Par & 1 & 0.8243 & 0.8248 & 0.7019 \\
& 10 & 0.8750 & 0.8063 & 0.6755 \\
& 20 & 0.8992 & 0.7615 & 0.6148 \\
& 30 & 0.9164 & 0.7532 & 0.6041 \\
\hline
Sat & 1 & 0.8105 & 0.8254 & 0.7027 \\
& 10 & 0.8095 & 0.8258 & 0.7033 \\
& 20 & 0.8118 & 0.8258 & 0.7033 \\
& 30 & 0.8108 & \bf 0.8261 & \bf 0.7037 \\
\hline
\end{tabular}
\end{table}

\begin{figure}[!t]
\centering
\includegraphics[width=0.8\linewidth]{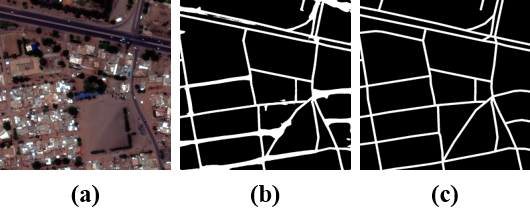}
\caption{An illustration of the effects of the MP loss (Both, $\lambda = 30$). (a) Satellite image; (b) Generated road map (from the partial branch); (c) Complete road map.}
\label{mp_sample}
\end{figure}

\begin{figure}[!t]
\centering
\includegraphics[width=1\linewidth]{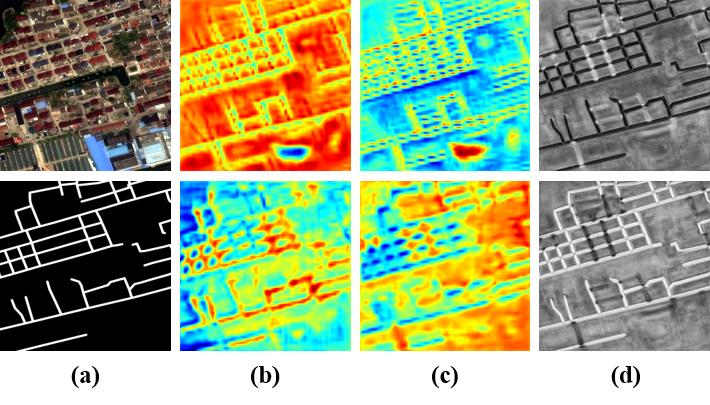}
\caption{Visualization of Gate Mechanism. The upper row is related to satellite branch, while the bottom row is related to partial branch. (a) Inputs; (b) $5^{th}$ channel maps from $F_S^{\prime}$ and $F_P^{\prime}$; (c) $27^{th}$ channel maps from $F_S^{\prime}$ and $F_P^{\prime}$; (d) Gate values.}
\label{feature_maps}
\end{figure}

The results are shown in Table \ref{mp_loss_strategy}, where the $R$ scores are also included since the direct target of introducing the MP loss is to improve $R$ scores. We observe that Both and Par have similar patterns. With the increase of $\lambda$, the $R$ scores are improved greatly, but at the cost of wrongly classifying many background pixels as road pixels. 
An example is shown in Figure \ref{mp_sample} for illustration.

This phenomenon is caused by adding the MP loss to the main branch, i.e. the partial branch. The purpose of the MP loss is to give more attention to the missing parts, which would have a weight of $1 + \lambda$, while other pixels have a weight of $1$. Therefore, once $\lambda$ is set too large, the model would focus on the missing parts and neglect other pixels (as  their weights are relatively small). 
As for Sat, the outputs of it would suffer from the same problem, but its feature maps would contain more coarse and richer information, which could be filtered during fusion to produce better outputs in partial branch. From Table \ref{mp_loss_strategy},
Sat appears to be more stable, and the scores of $F1$ and $IoU$ are the highest when $\lambda = 30$. 

\subsubsection{Efficacy of GSAM and MP Loss}

\noindent Our P2CNet is built on the two-branch Features Fusion model, with the MP loss added to the satellite branch only (Sat). P2CNet has two prominent components: GSAM and MP loss. We study their effects by incorporating them to the basic model gradually and report on $F1$ and $IoU$ metrics. The results are shown in Table \ref{ablation}, where all models are trained and tested on mix dataset.

\begin{table}[!t]
\caption{Ablation study on P2CNet (mix dataset, SpaceNet).}
\label{ablation}
\centering
\begin{tabular}{lrr}
\hline
Method & $F1$ & $IoU$ \\
\hline
Basic & 0.8248 & 0.7019 \\
+ MP & 0.8261 & 0.7037 \\
+ MP + GSAM (Gate (exist)) & 0.8257 & 0.7032 \\
+ MP + GSAM (Gate) & 0.8266 & 0.7045 \\
+ MP + GSAM (Gate + Attention (exist) & 0.8272 & 0.7053 \\
+ MP + GSAM (Gate + Attention) & \bf 0.8285 & \bf 0.7071 \\
\hline
\end{tabular}
\end{table}

\begin{table*}[!t]
\caption{Comparisons of models for road extraction. All models are trained and tested on 0\% SpaceNet dataset; The table is sorted based on $mIoU$ in ascending order. (par) means outputs from partial branch.}
\label{robust_no_partial_spacenet}
\centering
\resizebox{.6\linewidth}{!}{
    \begin{tabular}{lrrrrrr}
    \hline
    \multirow{2}{*}{Method} & \multicolumn{6}{c}{SpaceNet} \\
    & $P$ & $R$ & $F1$ & $APLS$ & $IoU$ & $mIoU$ \\
    \hline
    CasNet~\cite{cheng2017automatic} & 0.5991 & 0.6561 & 0.6263 & 0.4266 & 0.4559 & 0.7028 \\
    PSPNet~\cite{zhao2017pyramid} & 0.6330 & 0.6512 & 0.6420 & 0.4825 & 0.4727 & 0.7130 \\
    SGCN~\cite{zhou2021split} & 0.6344 & 0.6940 & 0.6629 & 0.5557 & 0.4957 & 0.7251 \\
    DLinkNet~\cite{zhou2018d} & 0.6530 & 0.6721 & 0.6624 & 0.5743 & 0.4953 & 0.7256 \\
    LinkNet~\cite{chaurasia2017linknet} & 0.6340 & 0.7013 & 0.6659 & 0.5588 & 0.4992 & 0.7269 \\
    DeepLabV3~\cite{chen2017rethinking} & 0.6621 & 0.6794 & 0.6706 & 0.5807 & 0.5045 & 0.7307 \\
    RoadCon~\cite{batra2019improved} & 0.6763 & 0.6705 & 0.6734 & 0.5719 & 0.5076 & 0.7329 \\
    PAN~\cite{li2018pyramid} & 0.6747 & 0.6913 & 0.6829 & 0.5986 & 0.5185 & 0.7386 \\
    FPN~\cite{kirillov2017unified} & 0.6787 & 0.6875 & 0.6831 & 0.6044 & 0.5187 & 0.7388 \\
    DeepLabV3+~\cite{chen2018encoder} & 0.6717 & \bf 0.7112 & 0.6909 & \bf 0.6192 & 0.5277 & 0.7433 \\
    \hline
    P2CNet (ours, par) & \bf 0.6944 & 0.6902 & \bf 0.6923 & 0.6061 & \bf 0.5297 & \bf 0.7451 \\
    \hline
    \end{tabular}
}
\end{table*}

\begin{table*}[!t]
\caption{Comparisons of models for road extraction. All models are trained and tested on 0\% OSM dataset; The table is sorted based on $mIoU$ in ascending order. (par) means outputs from partial branch.}
\label{robust_no_partial_osm}
\centering
\resizebox{.6\linewidth}{!}{
    \begin{tabular}{lrrrrrr}
    \hline
    \multirow{2}{*}{Method} & \multicolumn{6}{c}{OSM} \\
    & $P$ & $R$ & $F1$ & $APLS$ & $IoU$ & $mIoU$ \\
    \hline
    PSPNet~\cite{zhao2017pyramid} & 0.6317 & 0.6559 & 0.6436 & 0.5245 & 0.4745 & 0.7039 \\
    PAN~\cite{li2018pyramid} & 0.6425 & 0.6882 & 0.6481 & 0.5874 & 0.4794 & 0.7053 \\
    CasNet~\cite{cheng2017automatic} & 0.6226 & 0.6888 & 0.6540 & 0.5701 & 0.4859 & 0.7094 \\
    DeepLabV3~\cite{chen2017rethinking} & 0.6463 & 0.6975 & 0.6709 & 0.6334 & 0.5048 & 0.7209 \\
    RoadCon~\cite{batra2019improved} & 0.6679 & 0.6871 & 0.6774 & \bf 0.6723 & 0.5121 & 0.7259 \\
    SGCN~\cite{zhou2021split} & 0.6644 & 0.7009 & 0.6821 & 0.6485 & 0.5176 & 0.7287 \\
    FPN~\cite{kirillov2017unified} & 0.6767 & 0.6892 & 0.6829 & 0.6468 & 0.5184 & 0.7297 \\
    DeepLabV3+~\cite{chen2018encoder} & 0.6573 & \bf 0.7147 & 0.6848 & 0.6639 & 0.5207 & 0.7300 \\
    LinkNet~\cite{chaurasia2017linknet} & 0.6735 & 0.7015 & 0.6872 & 0.6678 & 0.5235 & 0.7323 \\
    DLinkNet~\cite{zhou2018d} & 0.6684 & 0.7080 & 0.6876 & 0.6575 & 0.5240 & 0.7323 \\
    \hline
    P2CNet (ours, par) & \bf 0.6899 & 0.6860 & \bf 0.6879 & 0.6613 & \bf 0.5243 & \bf 0.7335 \\
    \hline
    \end{tabular}
}
\end{table*}

We first employ MP loss to the basic model, and the $IoU$ score is improved from 70.19\% to 70.37\%. Then, we study our proposed GSAM which consists of an attention and a gate mechanism. 
We highlight the differences among our designed modules and existing ones as follow. {\bf (a)} For gate mechanism, the shape of $F_C^{\prime}$ in Figure \ref{gsam} is $2 \times C \times H \times W$, but that of existing works is $2 \times H \times W$, which means we would have $C$ confidence maps for each pair of feature maps of $F_S^{\prime}$ and $F_P^{\prime}$ for each channel, while existing work would uniformly have one confidence map for all channels. To show the effect of such design, we conduct an experiment named ({\bf + MP + GSAM (Gate (exist))}), in which a shared confidence map is used for all channels, and the $IoU$ score is 70.32\% and it is 0.13\% less than that of ({\bf + MP + GSAM (Gate)}). {\bf (b)} And for attention mechanism, the one that does not use fused affinity matrix ({\bf + MP + GSAM (Gate + Attention (exist))}) would result in an $IoU$ score of 70.53\%, which is 0.18\% less than the model with it ({\bf + MP + GSAM (Gate + Attention)}), and this final model could achieve the best performance of 82.85\%, 70.71\% for $F1$ and $IoU$, respectively.

We also visualize some feature maps of the gate mechanism in GSAM in Figure \ref{feature_maps}, where in column (b) and (c), red values represent large values, while blue values are smaller ones.
As there are hundreds of channels per group of feature maps in the gate, we select the $5^{th}$ and $27^{th}$ channels of $F_S^{\prime}$ and $F_P^{\prime}$ to indicate that our proposed gate mechanism could highlight clear semantic areas, e.g. road and background in this case. In addition, the figures in column (d) are from two channels of $F_C^{\prime}$, which could fuse features from satellite and partial branches with different confidences in a complementary manner.

As we could observe from the second row, the feature map of column (c) has larger values on the ``black'' part of column (a), which means this feature map focus on extracting features from the useless part. Such information would be redundant during fusion in GSAM, and by utilizing the channel-wise self-attention, the confusable information could be safely filtered, and the fusion process could work better.

\subsubsection{Robustness Against No Partial Road Maps}

\noindent To show the robustness of different models, we also train all models except for McGAN on the 0\% dataset and then test them on the 0\% dataset. The reason behind is that in platforms like OSM, there are also many places that have few or even no partial road maps, and our proposed P2CNet is expected to behave well on such case.

We did not train the McGAN model, because its discriminators require the partial road maps as the inputs of noisy road maps. As our P2CNet is a two-branch model, we use all-zero maps as the inputs of the partial branch, while other models follow their original structure, i.e., they act as normal road extraction models, and do not use partial road maps.
For our P2CNet, the missing parts of MP loss are road pixels in complete road maps, and the weights of road pixels appear to be twice as large as those of background pixels.
The results are shown in Table~\ref{robust_no_partial_spacenet} and Table~\ref{robust_no_partial_osm},
and show that P2CNet is still slightly better than the state-of-the-art DeepLabV3+ (0.5297 vs 0.5277) on SpaceNet and DLinkNet (0.5243 vs 0.5240) on OSM for $IoU$.

\section{Conclusion}
\label{sec:conclusion}

\noindent In this paper, we study the road extraction task based on satellite images and partial road maps. We introduce a two-branch framework called Partial to Complete Network (P2CNet) for the task. P2CNet involves two prominent components: Gated Self-Attention Module (GSAM) and MP loss. GSAM fuses the feature maps from two branches and MP loss helps to put more attention to missing road pixels. Extensive comparative experiments were conducted, which show that P2CNet achieves the best performance compared with existing state-of-the-art models. 
One interesting research direction is to use publicly available richer data such as different layers of OSM for other related tasks such as semantic map labelling.
Moreover, the strategy could be further improved by taking road types into consideration, which could be studied in future work.

\bibliographystyle{IEEEtran}
\bibliography{ref.bib}

\vskip -2\baselineskip plus -1fil
\begin{IEEEbiography}[{\includegraphics
[width=1in,height=1.25in,clip,
keepaspectratio]{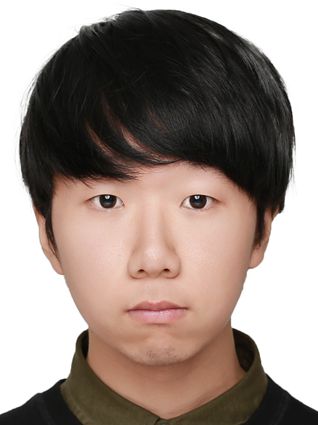}}]
{Qianxiong Xu} 
is currently pursuing his Ph.D. degree in the School of Computer Science and Engineering (SCSE), Nanyang Technological University (NTU). He received the Bachelor degree from Nanjing University of Information Science and Technology in 2020. His research interests include data mining, remote sensing, and computer vision.
\end{IEEEbiography}
\vskip -2\baselineskip plus -1fil
\begin{IEEEbiography}[{\includegraphics
[width=1in,height=1.25in,clip,
keepaspectratio]{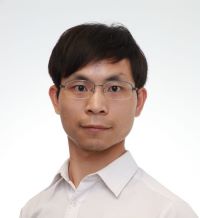}}]
{Cheng Long} 
is currently an Assistant Professor at the School of Computer Science and Engineering (SCSE), Nanyang Technological University (NTU). From 2016 to 2018, he worked at Queen’s University Belfast, UK. He got the PhD degree from the Department of Computer Science and Engineering, The Hong Kong University of Science and Technology (HKUST) in 2015. His research interests include data management, data mining and big data analytics.
\end{IEEEbiography}
\vskip -2\baselineskip plus -1fil
\begin{IEEEbiography}[{\includegraphics
[width=1in,height=1.25in,clip,
keepaspectratio]{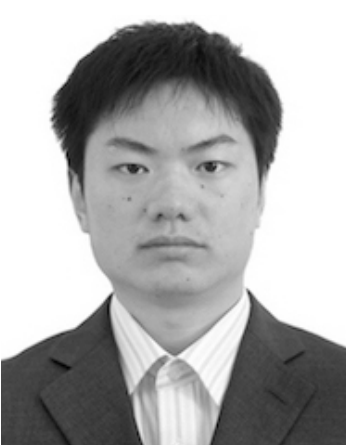}}]
{Liang Yu} 
received the Ph.D. degree in photogrammetry and remote sensing from Wuhan University in 2008. He did the Post-Doctoral Research in the National University of Singapore, the National Center for Supercomputing Applications, University of Illinois at Urbana–Champaign, and the Singapore-MIT Alliance for Research and Technology. He had been leading the Smart Mobility Group, Institute of Infocomm Research, Singapore, where he initiated many research efforts to transportation and urban planning. He is currently a Senior Data Scientist with Alibaba Cloud, where he is leading the algorithm development for the City Brain Project. His research interest has covered a wide range of smart city topics such as geospatial data integration, data semantics, and data analytics.
\end{IEEEbiography}
\vskip -2\baselineskip plus -1fil
\begin{IEEEbiography}[{\includegraphics
[width=1in,height=1.25in,clip,
keepaspectratio]{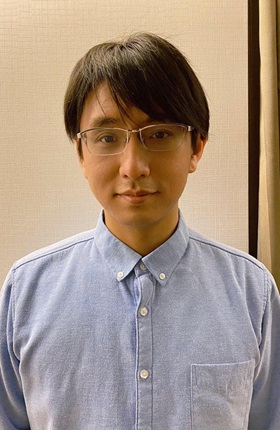}}]
{Chen Zhang} 
is currently a Research Assistant Professor of the Department of Computing, The Hong Kong Polytechnic University (PolyU). Before joining the Department, he worked as a senior manager of the Big Data Institute at The Hong Kong University of Science and Technology (HKUST). He received his Ph.D. degree in Computer Science and Engineering from HKUST in 2015, supervised by Prof.Lei CHEN. Dr. Zhang is broadly interested in Crowdsourcing, Fintech, and Machine Learning.
\end{IEEEbiography}
\vfill

\clearpage

\section*{Appendix}
\label{sec:appendix}

\begin{figure}[h]
    \centering
    \includegraphics[width=\linewidth]{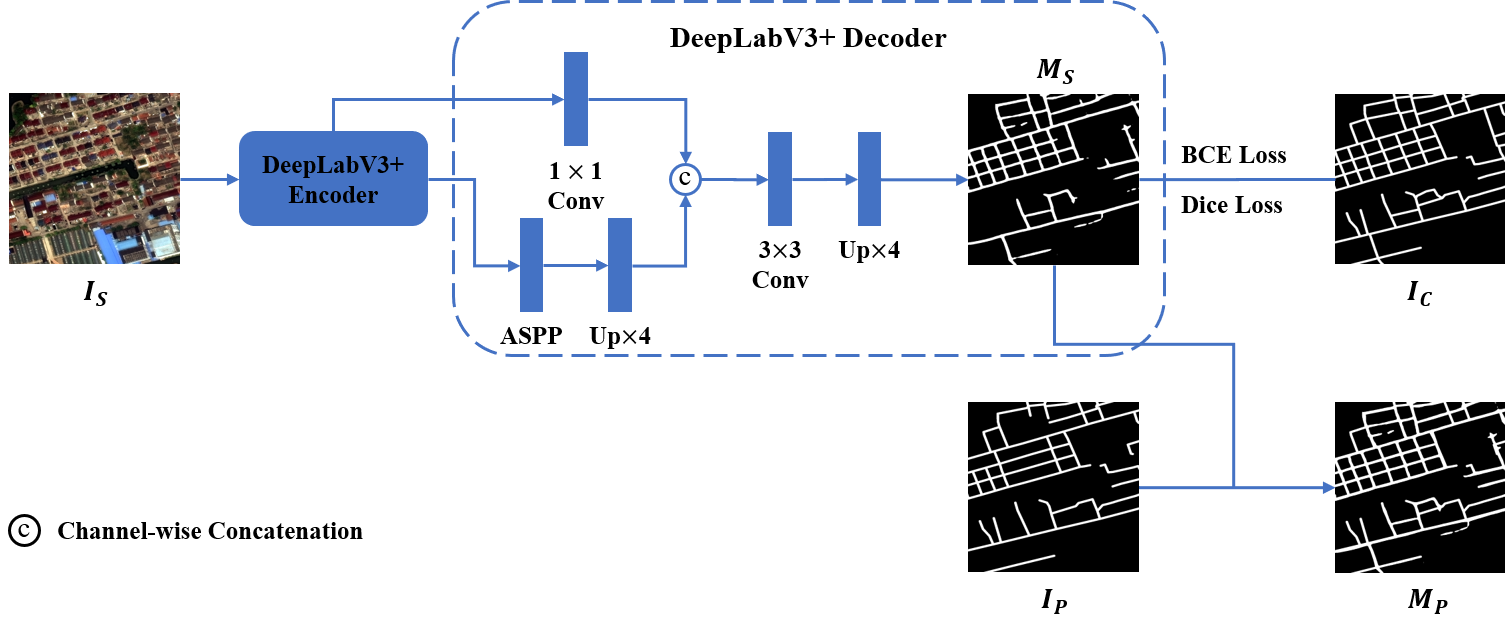}
    \caption{Model Structure of Outputs Augmentation.}
    \label{outputs_augmentation}
\end{figure}

\begin{figure}[h]
    \centering
    \includegraphics[width=\linewidth]{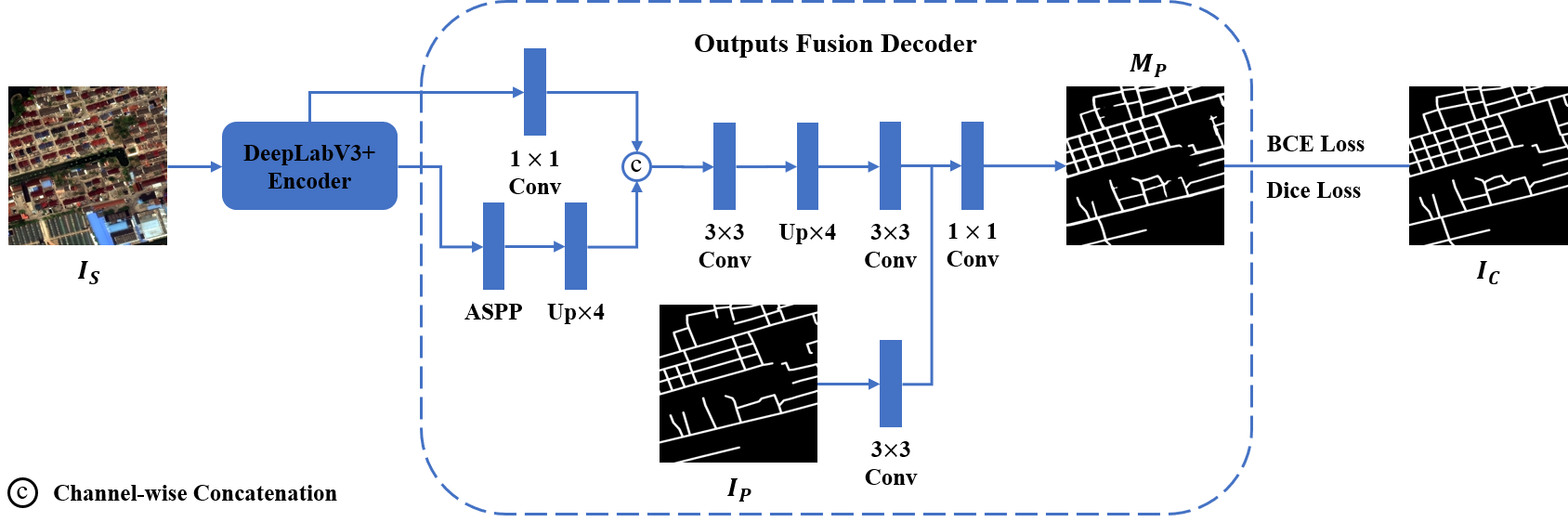}
    \caption{Model Structure of Outputs Fusion.}
    \label{outputs_fusion}
\end{figure}

\begin{figure}[h]
    \centering
    \includegraphics[width=\linewidth]{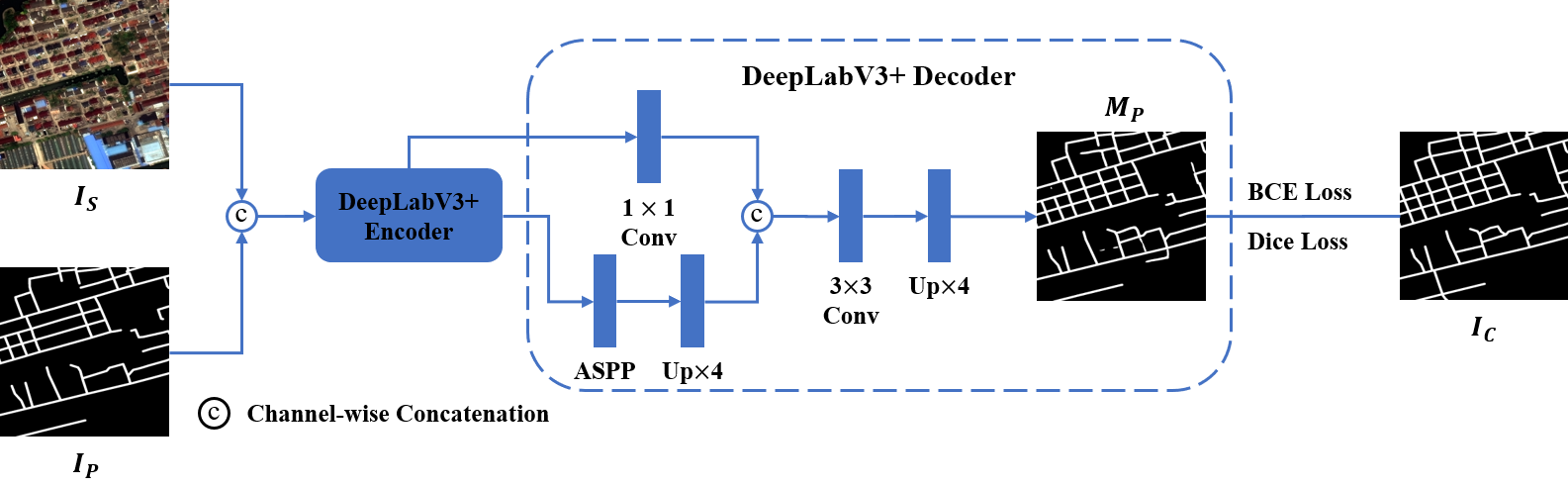}
    \caption{Model Structure of Inputs Fusion.}
    \label{inputs_fusion}
\end{figure}

\begin{figure}[h]
    \centering
    \includegraphics[width=\linewidth]{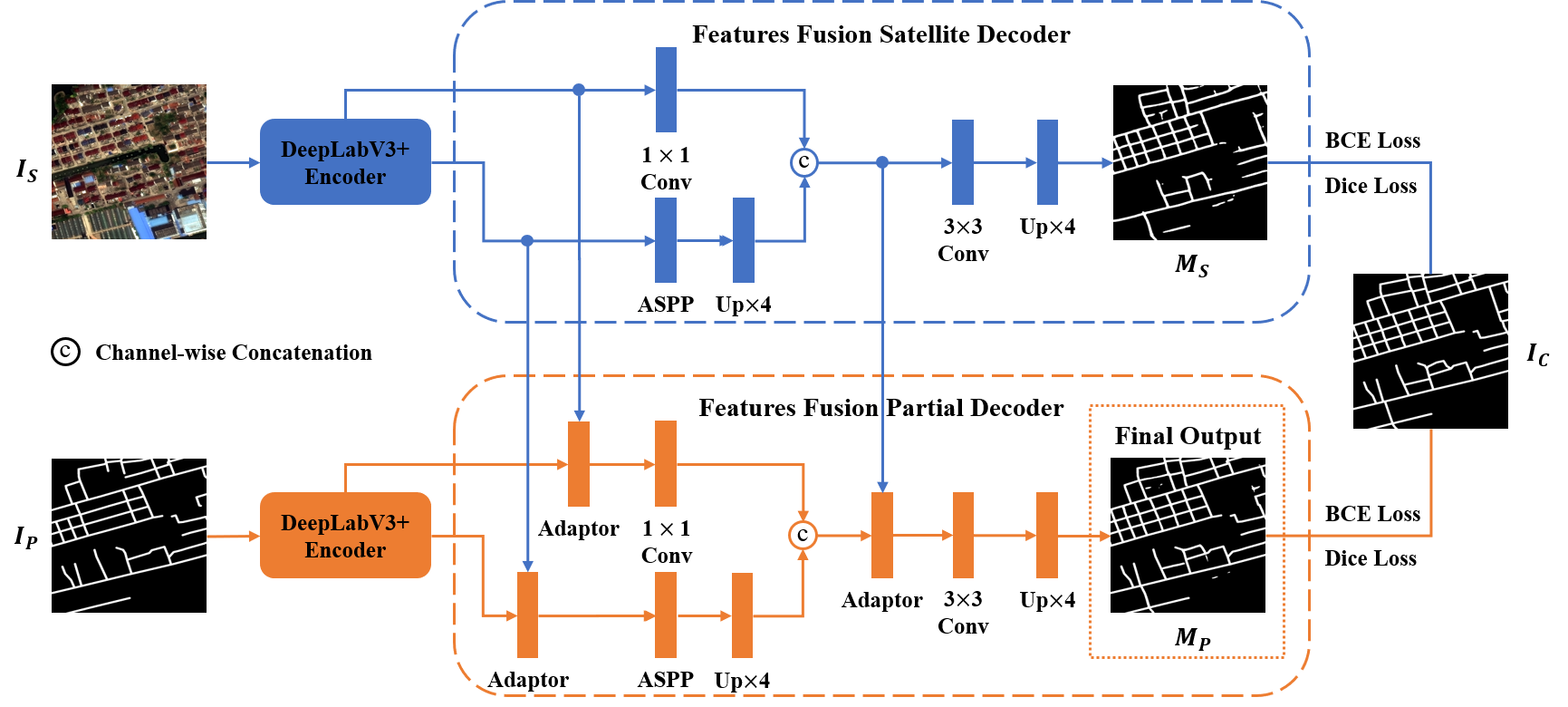}
    \caption{Model Structure of Features Fusion.}
    \label{features_fusion}
\end{figure}

\noindent Four basic fusion methods that could be applied to our setting, say, fusing satellite images and partial road maps, are briefly introduced as follows. Note that we show the models with a uniform basic model, DeepLabV3+~\cite{chen2018encoder}.
\medskip\\
{\bf Outputs Augmentation:} Figure \ref{outputs_augmentation} shows the model structure of Outputs Augmentation. In this case, the partial road maps are fused in a coarse way, and it does not require any extra parameters. Specifically, we train a basic DeepLabV3+, with the backbone of ResNet-34, then take an average of the binarized outputs of DeepLabV3+ and partial road maps to obtain the fused outputs.
\medskip\\
{\bf Outputs Fusion:} As Figure \ref{outputs_fusion} illustrates, there are three more convolution layers than DeepLabV3+, where the two $3 \times 3$ convolution layers are used to convert the inactivated predictions and partial road maps into the same space, and another $1 \times 1$ convolution layer is used to fuse them.
\medskip\\
{\bf Inputs Fusion:} The only difference between the basic DeepLabV3+ and the model of Inputs Fusion lies in the first convolution layers, where their numbers of input channels are 3 and 4, respectively. In Inputs Fusion, the satellite images and corresponding partial road maps are concatenated along channel to be the inputs. The partial road maps would influence the ability of extracting features significantly, as they are binary images and would be regarded as a kind of strong condition, using which the model would pay more attention on reserving existing roads. The model of Inputs Fusion is shown in Figure \ref{inputs_fusion}.
\medskip\\
{\bf Features Fusion:} 
The number of parameters of Features Fusion is more than twice as large as that of DeepLabV3+. As Figure \ref{features_fusion} shows, the middle-layer low-level features are fused together to generate better results. In this case, it is clear that there are two branches, one for satellite images, while another for partial road maps, and they could be jointly trained by two sets of loss functions. The satellite branch act as a dedicate feature extractor, while the partial branch aims to not only extract useful information from partial road maps, but also fuse the information of satellite images and partial road maps, to yield better outputs.

\end{document}